\documentclass[conference]{IEEEtran}
\IEEEoverridecommandlockouts
\usepackage{cite}
\usepackage{amsmath,amssymb,amsfonts}
\usepackage{algorithmic}
\usepackage{graphicx}
\usepackage{textcomp}
\usepackage{xcolor}
\def\BibTeX{{\rm B\kern-.05em{\sc i\kern-.025em b}\kern-.08em
    T\kern-.1667em\lower.7ex\hbox{E}\kern-.125emX}}

\usepackage{wrapfig}
\usepackage{amsmath}

\usepackage{amssymb}
\usepackage{mathtools}
\usepackage{amsthm}
\usepackage{booktabs}
\usepackage[switch]{lineno}
\usepackage{colortbl}
\usepackage{multirow}
\usepackage[ruled]{algorithm2e}
\definecolor{bluex}{RGB}{54, 125, 189}
\usepackage[colorlinks=True,
            linkcolor=red,
            anchorcolor=blue,  
            pagebackref,  
            urlcolor=black,
            citecolor=bluex,
            ]{hyperref}

\usepackage{diagbox}
\usepackage{color}
\usepackage{xcolor}
\usepackage{algorithmic}
\usepackage{alltt}
\usepackage{pifont}
\usepackage{subcaption}

\usepackage{array}
\usepackage{makecell}
\usepackage{rotating}
\usepackage{caption}
\usepackage{float}
\usepackage{afterpage}
\usepackage{stfloats}

\definecolor{ForestGreen}{rgb}{0.13, 0.55, 0.13}
\definecolor{Black}{rgb}{0,0,0}

\definecolor{mygray}{gray}{.9}

\usepackage{colortbl} % To add color to v-lines
\usepackage{graphicx} % To resize table
\usepackage{enumitem}  % fancy enumerate
\usepackage{xspace}  % for adding space for 

\newcommand{\eevpt}[0]{E$^{2}$VPT}

\newcommand{\vtab}[0]{VTAB-1k}

\definecolor{tabvline}{HTML}{a8a495}
\definecolor{prompt_blue}{HTML}{1f78b4}
\definecolor{prompt_red}{HTML}{d45c43}

\definecolor{green_im}{rgb}{0.0, 0.5, 0.0}
\usepackage{amsmath,amsfonts,bm}

\def\eqref#1{equation~\ref{#1}}
\def\1{\bm{1}}

\def\vp{{\bm{p}}}

\def\vv{{\bm{v}}}

\def\vx{{\bm{x}}}

\def\vz{{\bm{z}}}

\DeclareMathAlphabet{\mathsfit}{\encodingdefault}{\sfdefault}{m}{sl}
\SetMathAlphabet{\mathsfit}{bold}{\encodingdefault}{\sfdefault}{bx}{n}

\def\ninept{\def\baselinestretch{.95}\let\normalsize\small\normalsize}

\begin{document}
\ninept

% \title{Conference Paper Title*\\
% {\footnotesize \textsuperscript{*}Note: Sub-titles are not captured for https://ieeexplore.ieee.org  and
% should not be used}
% \thanks{Identify applicable funding agency here. If none, delete this.}
% }

\title{Semantic Hierarchical Prompt Tuning for Parameter-Efficient Fine-Tuning}

\author{%
 \IEEEauthorblockN{Haowei Zhu, Fangyuan Zhang, Rui Qin, Tianxiang Pan, Junhai Yong, Bin Wang$^\dag$}  
\IEEEauthorblockA{\textit{School of Software} \\
\textit{Tsinghua University}\\
Beijing, China \\
$^\dag$wangbins@tsinghua.edu.cn}
  % \\\texttt{~~wangbins@tsinghua.edu.cn}\\
}

\maketitle

\begin{abstract}
As the scale of vision models continues to grow, Visual Prompt Tuning (VPT) has emerged as a parameter-efficient transfer learning technique, noted for its superior performance compared to full fine-tuning. However, indiscriminately applying prompts to every layer without considering their inherent correlations, can cause significant disturbances, leading to suboptimal transferability. Additionally, VPT disrupts the original self-attention structure, affecting the aggregation of visual features, and lacks a mechanism for explicitly mining discriminative visual features, which are crucial for classification.
To address these issues, we propose a Semantic Hierarchical Prompt (SHIP) fine-tuning strategy. We adaptively construct semantic hierarchies and use semantic-independent and semantic-shared prompts to learn hierarchical representations. We also integrate attribute prompts and a prompt matching loss to enhance feature discrimination and employ decoupled attention for robustness and reduced inference costs. SHIP significantly improves performance, achieving a 4.9\% gain in accuracy over VPT with a ViT-B/16 backbone on \vtab{} tasks. Our code is available at \href{https://github.com/haoweiz23/SHIP}{https://github.com/haoweiz23/SHIP}.

\end{abstract}

\begin{IEEEkeywords}
semantic hierarchical prompt, parameter efficient fine-tuning
\end{IEEEkeywords}

\section{Introduction}
\label{sec:intro}

The rise of artificial intelligence (AI) has increased the reliance on large-scale pre-trained foundation models \cite{vit, CLIP, stablediffusion}, particularly for visual tasks, such as image classification \cite{vit, zhu2022dual} and generation \cite{distdiff, dipgo}. Traditionally, adapting these models to specific tasks involves full fine-tuning, which is inefficient due to high storage and computation costs, as it requires storing and learning all the parameters of the model for specific tasks and deployment scenarios \cite{jia2022vpt}.

Parameter-Efficient Fine-Tuning (PEFT) methods reduce storage and computational demands by adjusting only a subset of the parameters of a model while keeping the rest frozen, has emerged as a viable alternative \cite{jia2022vpt, he2023sensitivity, zhang2022neural, cheng2023e2vpt, jie2022convolutional, chen2022adaptformer}. However, existing PEFT methods, such as extra modules \cite{hu2022lora, zhang2020side, chen2022adaptformer, zhao2023sct}, prompt tuning \cite{jia2022vpt, cheng2023e2vpt, pei2024sa2vp, wang2024lion}, and compound tuning \cite{zhang2022neural, he2023sensitivity}, often apply uniform parameters across tasks and models, which may not fully address the custom needs of specific tasks. Besides, existing methods introduce additional prompts into the embedded token sequence, which can disrupt the original attention structure. Furthermore, these methods lack an explicit mechanism for mining discriminative visual clues.

\begin{figure}[htbp]
    \centering
    \includegraphics[width=\linewidth]{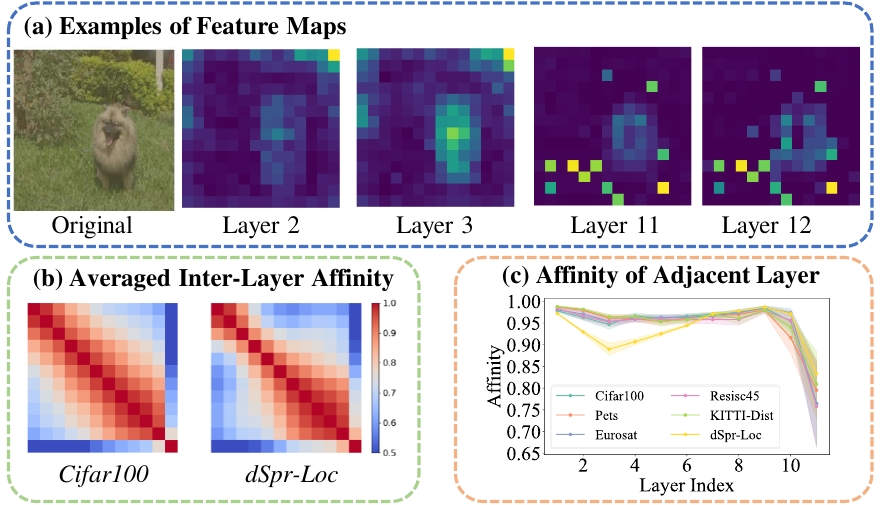}
    % \vspace{-5pt}
    \caption{(a) Examples of feature maps in the continuous transformer layers of ViT-B/16 \cite{vit}. Semantic patterns exhibit similarity between adjacent pairs of layers and gradually evolve in deeper layers. (b) The averaged inter-layer affinity matrix is computed across all training samples from three typical datasets. (c) The statistical mean and variance of the affinity between the $i$-th and $(i+1)$-th layers are analyzed across six typical datasets.}
    \label{fig:observation}
\end{figure}

To address these limitations, we introduce the Semantic Hierarchical Prompt (SHIP) fine-tuning method, which identifies semantic hierarchies within pre-trained models and assigns trainable prompts based on semantic correlations. 
We analyze layer-wise feature characteristics (see Fig.~\ref{fig:observation}), revealing that adjacent layers in Vision Transformers (ViT) \cite{vit} exhibit high feature similarity, while semantic features vary significantly across tasks. For instance, object-level tasks show notable feature transitions in later layers, whereas non-object-level tasks demonstrate early-stage shifts. This observation motivates our approach: different tasks possess distinct semantic hierarchies and therefore require semantic-specific prompts.

Based on these insights, we propose an adaptive strategy to determine the semantic granularity of pre-trained models for downstream tasks. Specifically, we extract intermediate features from transformer layers, construct an inter-layer affinity matrix, and determine semantic hierarchies with threshold-based greedy search. We then introduce Semantic-Independent Prompts (SIP) for different semantic hierarchies to learn semantic-specific features.
Additionally, we develop Semantic-Shared Prompts (SSP) to provide shared representations across layers, aligning features among transformer layers and mining common inter-layer characteristics. Unlike VPT-Deep or VPT-Shallow \cite{jia2022vpt}, SSP inputs the same values across all layers.
Furthermore, prior research has shown that attribute features like color and shape serve as effective prompts for downstream tasks \cite{tian2024argue, kim2024aapl}. Building on this, we introduce Attribute Prompts (AP) to aggregate attribute patterns from training data. Specifically, we construct an attribute prototype set by clustering the training subset into groups and averaging their features. High-response image tokens are then used to aggregate attribute-related features, which, combined with a set of trainable parameters, form the Attribute Prompt (AP).
To explicitly mine discriminative visual clues, we propose a Prompt Matching Loss (PML) to align prompts with high-response instance tokens. Moreover, we introduce a Decoupled Attention (DA) mechanism that separates the vanilla attention into three attention components, preserving the original instance attention structure and enhancing training robustness.
Our approach is both effective and easy to implement, offering significant improvements over existing methods with minimal parameter usage. For example, SHIP achieves a 4.9\% increase in average Top-1 accuracy over VPT on the \vtab{} dataset while using only 0.38M trainable parameters.

\section{Related Work}
% \subsection{Parameter-Efficient Fine-tuning of Visual Pretrained Models}
\label{sec: peft}
%Full fine-tuning is a direct and effective method for applying large pre-trained models to downstream tasks. However, as the scale of the model increases, the associated training and storage burdens become prohibitive. Parameter-Efficient Fine-Tuning (PEFT) offers an efficient alternative by adjusting only a minimal set of parameters, thereby reducing storage requirements. Currently, parameter-efficient fine-tuning methods can be broadly categorized into three types: extra module methods, prompt tuning, and compound tuning methods.
Full fine-tuning incurs significant training and storage costs as models grow. Parameter-Efficient Fine-Tuning (PEFT) mitigates this by adjusting only a small subset of parameters, including the extra module, prompt tuning, and compound tuning methods.

\textbf{\textit{Extra module methods}} add trainable modules to the model while keeping most pre-trained parameters frozen \cite{cai2020tinytl}. Examples include Adapter \cite{houlsby2019parameter, zhang2021tip, chen2022adaptformer, jie2023revisiting, rebuffi2017learning, pfeiffer2020adapterhub, wang2024caps}, LoRA \cite{hu2022lora}, and Side \cite{zhang2020side}, which use additional structures to modulate or bypass parts of the network. The recent LION method \cite{wang2024lion} introduces two equilibrium implicit layers at both ends of the pre-trained backbone, improving flexibility and reducing memory usage across architectures.

\textbf{\textit{Prompt tuning methods}} introduce token-level \cite{jia2022vpt, cheng2023e2vpt, dapt, bahng2022exploring, chen2023understanding, tsao2024autovp, xu2023progressive, lin2023hierarchical} or pixel-level \cite{visual_pixel, Huang_2023_CVPR} learnable prompts into the input space without modifying pretrained parameters, thereby gaining popularity in large-scale vision backbone transfer learning. Besides, prompt-based methods only modify the input space, providing superior universality and scalability. Our method falls within this paradigm. 

\textbf{\textit{Compound tuning methods}} \cite{chowdhury2023apollo, he2022dynamic, shah2023adept} combine different techniques, establishing mechanisms between prompts and adapters to boost performance. Although promising, these methods may not fully account for the semantic layer relationships, often adding parameters without discrimination across the network and tasks.

Recent advancements leverage search algorithms \cite{zhang2022neural}, pruning \cite{cheng2023e2vpt}, and parameter sensitivity analysis \cite{he2023sensitivity} to dynamically optimize model architectures for specific tasks. Our method stands out by intuitively identifying semantic hierarchies within the model and generating tailored prompt tokens based on inter-layer correlation strengths. By organically integrating semantic-specific prompting strategies, it enhances feature aggregation across semantic levels, resulting in superior fine-tuning performance compared to other PEFT methods.

\begin{figure*}[htbp]
  \centering
    \includegraphics[width=\linewidth]{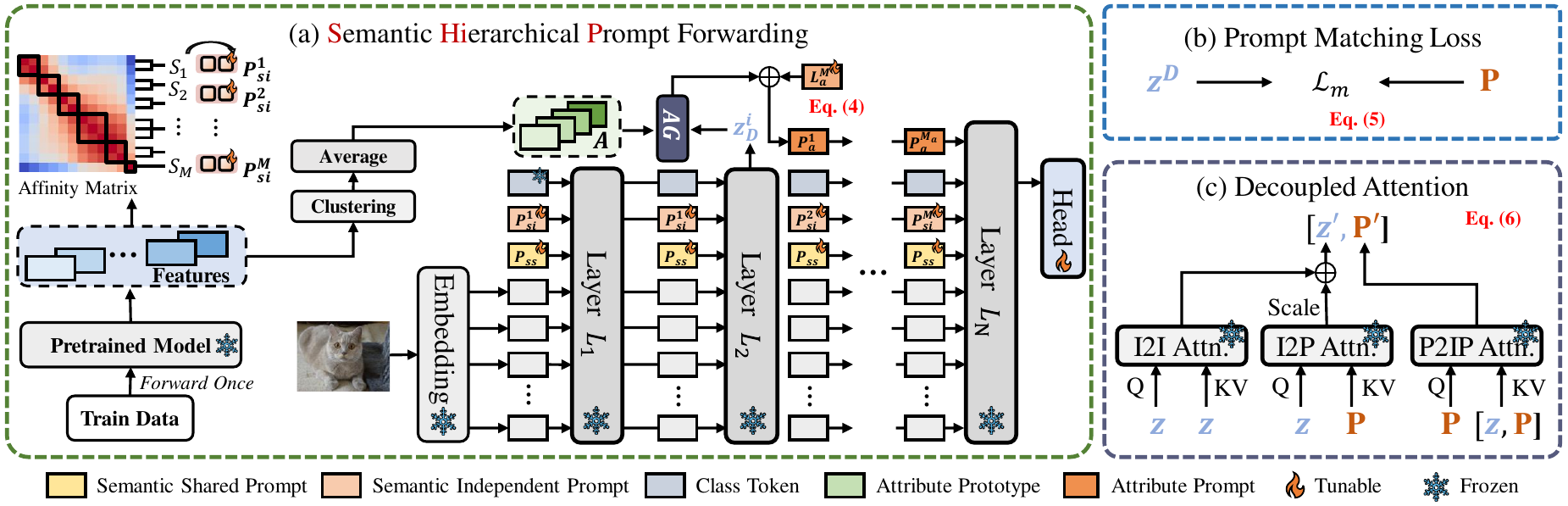}
  \caption{Overview of our proposed SHIP (\textbf{S}emantic \textbf{HI}erarchical \textbf{P}rompt) fine-tuning framework. SHIP computes inter-layer affinity using features derived from the pre-trained model. Based on this affinity, a semantic hierarchy is established through greedy search. SHIP then learns specific prompts for each semantic level, integrating them with prompt matching loss and decoupled attention to enhance model performance.}
  % \vspace{-10pt}
  \label{fig:overview}
\end{figure*}

\section{Method}
\subsection{Revisit Visual Prompt Tuning}\label{sec:method}
VPT \cite{jia2022vpt} is a pioneering work that applies prompt learning to visual tasks. It fine-tuned learnable prompt tokens into the input space of ViT while freezing the entire backbone. VPT has two variants: Shallow and Deep. VPT-Shallow only introduces prompts at the beginning of the model. Formally, denoting a collection of $N_p$ trainable prompts with $d$-dimension as $\mathbf{P}=\{\vp^i \in \mathbb{R}^d \mid i=1,2,\ldots, N_p\}$, VPT-Shallow is defined as follows:
\begin{equation}\label{Eq:shallow_vpt}
    \left[\vz_1, \vv_1 \right] =  L_1(\left[\vx_0, \mathbf{P}]\right]),
\end{equation}
\begin{equation}\label{Eq:shallow_vpt_2}
    \left[\vz_i, \vv_i \right] = L_i(\left[\vz_{i-1}, \vv_{i-1}\right]),
\end{equation}
where $\vz_i \in \mathbb{R}^{N_x \times d}$ and $\vv_i \in \mathbb{R}^{N_p \times d}$ represent the instance and prompt features processed by layer $L_i$, respectively. $N_x$ is the instance sequence length. 

VPT-Deep introduces additional prompt tokens at each layer, possessing a larger fine-tuning capacity, and has been demonstrated to exhibit superior transfer performance. It is formulated as follows:
\begin{equation}\label{Eq:deep_vpt}
    \left[\vz_i, _, \vv_i \right] = L_i(\left[\vz_{i-1},  \mathbf{P}_{i-1}\right]).
\end{equation}
In VPT, prompts are either inserted only in the first layer (VPT-Shallow) or uniformly across all layers (VPT-Deep), each with its own advantages and drawbacks. VPT-Shallow captures continuous semantics by sharing prompts across layers, leading to good adaptability and generalization but has limited parameter capacity. VPT-Deep, meanwhile, offers a larger trainable parameter capacity but neglects inter-layer semantic connections, which can confuse optimization, make the model sensitive to hyperparameters, and potentially reduce pre-training effectiveness. Additionally, they lack a mechanism to explicitly extract discriminative semantic features, which is crucial for visual recognition tasks. This raises three key questions: (1) Where should additional parameters be inserted in the pre-trained model to increase learning capacity? (2) At which positions should prompts be shared to effectively capture continuous semantics? (3) How can prompts provide discriminative semantic features?

\subsection{Semantic Hierarchical Prompt Tuning}
To address these questions, we analyzed the inter-layer feature characteristics of the pre-trained backbone. As shown in Fig. \ref{fig:observation}, we found high similarity among features in adjacent layers, with this similarity decreasing as the distance between layers increases. A clear semantic hierarchy is observed between layers, which varies across different tasks (Fig. \ref{fig:observation} (c)). Inspired by these findings, we propose constructing task-specific semantic levels and designing prompts for specific semantic hierarchies. Our Semantic Hierarchical Prompt (SHIP) fine-tuning method combines three types of prompts: (1) Semantic Independent Prompts (SIP), independent between hierarchies but shared within each hierarchy to consolidate semantics; (2) Semantic Shared Prompts (SSP), to complement the common features shared across layers; and (3) Attribute Prompts (AP), to incorporate sample-specific attributes. A Prompt Matching Loss (PML) is proposed to align prompts with discriminative visual cues. Additionally, we use a Decoupled Attention (DA) mechanism to separate the prompt learning from the original attention process. An overview of SHIP is shown in Fig. \ref{fig:overview}, with each component described below.

\textbf{Semantic Independent Prompt.}
Different semantic levels represent various hierarchical features. Shallow semantic representations capture high-frequency details like texture and edges, while deeper representations convey abstract low-frequency features. To aggregate features within a semantic level, a simple method is to divide layers evenly (e.g., every three layers form one level) \cite{swin, wang2021pyramid}. However, because semantic hierarchies can vary across tasks, this naive strategy may be suboptimal.
% , as demonstrated in Section \ref{sec:ab}. 
Therefore, we propose using \textit{Inter-Layer Affinity} for a more refined hierarchical partition. For a sample input to a pretrained ViT with $N$ transformer layers, yielding $N$ intermediate features $\vz_i$, the affinity $S_{ij}$ between the $i$-th and $j$-th layers is computed as the dot product (cosine similarity) between normalized $\vz_i$ and $\vz_j$. We average the affinity matrices of sampled training images to obtain a robust inter-layer affinity matrix. A greedy search strategy then groups layers with high affinity into the same semantic hierarchy: starting from layer \(i = 0\), the next \(a\) layers with affinity above threshold \(\lambda\) are grouped into the same hierarchy. We then update \(i = i + a\) and repeat until all layers are assigned. For $M$ semantic levels $\{S_i\}_{i=1}^M$, SHIP inserts $M$ sets of prompts $\mathbf{P_{si}}$ at the beginning of each semantic level. Within each level, prompts are shared as shown in Eq. \ref{Eq:shallow_vpt_2} to extract intra-semantic hierarchy patterns.

\textbf{Semantic Share Prompt.} 
In addition to sharing weights within each semantic level,  we introduce a set of shared prompts, $\mathbf{P_{ss}} \in \mathbb{R}^{N_{SS} \times d}$ ($N_{SS} = 10$ by default), that are applied across all semantic levels. These shared prompts are inserted into each layer following Eq. \ref{Eq:deep_vpt}, allowing downstream task representations to be evenly integrated throughout the model. This approach helps to complement the common features shared across layers, refine the model's overall representation, and reduce the gap between the pretrained model and the downstream task.

\textbf{Attribute Prompt.}  
Attributes, such as colors, shapes, or object parts, are visual features that aid in image classification. We select a subset $D_t$ from the training data and use K-means clustering \cite{macqueen1967some} to divide it into $K$ (default 200) subsets. Each cluster shares similar patterns and attributes. We compute the average feature values for each cluster as attribute prototypes $\mathcal{A} = \{a_i\}_{i=1}^K$. In the model forward process, we use a sample-aware strategy to adaptively aggregate these attributes and add them to a set of learnable parameters to create attribute prompts $\mathbf{P_{a}}$. Specifically, we select $N_{a}$ instance tokens $\vz_D^i$ with the highest attention scores from the $\hat{\texttt{CLS}}$ token of $i$-th layer (where $N_{a}$ is set to 10, matching the length of $\mathbf{P_{a}}$) and construct the attribute prompts $\mathbf{P_{a}}$ as shown in Eq. \ref{Eq:ap}. These prompts are then inserted into the subsequent input sequence. We apply attribute prompts to the last $M_a$ (default 2) hierarchies.

\begin{align}\label{Eq:ap}
    \mathbf{P_{a}} &= (1-\lambda_a) L_a + \lambda_a \mathrm{AG}\left(\vz_D^i, \mathcal{A}\right) \nonumber \\
    &= (1-\lambda_a) L_a + \lambda_a f \left(\vz_D^i, \mathcal{A} \right) \mathcal{A},
\end{align}
% \begin{equation}\label{Eq:ap}
%     \mathbf{P_{a}} = (1-\lambda_a) L_a + \lambda_a \mathrm{AG}, \mathrm{AG}= f \left(\vz_D^i, \mathcal{A} \right) \mathcal{A}  ,
% \end{equation}
where $\mathrm{AG}(\cdot)$ denotes the attribute aggregation operation of querying sample-related attributes from $\mathcal{A}$ using $\vz_D^i$ and performing a weighted aggregation of these attributes. $L_a \in \mathbb{R}^{N_a \times D}$ represents a set of learnable parameters, and $f(\cdot)$ computes and normalizes the cosine similarity between instance tokens and attribute prototypes. $\lambda_a$ is a scaling factor, set to 0.1 by default.

\textbf{Prompt Matching Loss.} 
To explicitly constrain prompts to focus on visually discriminative features, we introduce Prompt Matching Loss (PML). This matches prompts with the most discriminative visual tokens and calculates the cosine distance loss. Specifically, we first select $N_m$ (default 10) instance tokens $\vz_D$ with the highest attention scores. We then compute the cosine distances between all prompts in the current layer and these discriminative tokens, and calculate the cosine loss for the nearest neighbor prompt-instance token pairs as shown in Eq. \ref{Eq:pml}. The final loss $\mathcal{L}$ is a combination of the prompt matching loss $\mathcal{L}_m$ and the classification loss $\mathcal{L}_c$, where $\mathcal{L}_m$ is weighted by a coefficient $\lambda_m$ (default: 0.5) and applied exclusively to the output of the last transformer layer.

\begin{equation}\label{Eq:pml}
    \mathcal{L}_m = \frac{1}{N_p} \sum_{i=1}^{N_p} \min_j \left( 1 - \frac{\vp_i \cdot \vz_j^D}{\|\vp_i\| \|\vz_j\|} \right)
\end{equation}

\textbf{Decoupled Attention.} 
Vanilla attention causes the instance tokens' attention weights to be dispersed by prompts, which disrupts the pretrained model’s ability to aggregate features and worsens with longer prompt sequences. We address this by reformulating attention into three attention parts: instance-to-prompt (I2P), instance-to-instance (I2I), and prompt-to-instance-and-prompt (P2IP). This approach maintains the original instance self-attention structure, and enhances training stability without additional computational cost.
\begin{equation}\label{Eq:wi}
    \vz' = (1 - \lambda_{d}) \mathrm{I2I}(\vz, \vz) + \lambda_{d}  \mathrm{I2P}(\vz, \mathbf{P}),
    \mathbf{P'} = \mathrm{P2IP}(\mathbf{P}, [\vz, \mathbf{P}]),
\end{equation}
where $\lambda_d$ is a hyperparameter with a default value of 0.1. The P2IP component aggregates instance features for prompts. However, we have found that it can be omitted with minimal impact on performance. Therefore, we exclude the P2IP component except for the layer introducing the prompt matching loss.

\section{Experiments}
\subsection{Setup}
\noindent\textbf{Setting.} 
Our experiments are conducted on the \vtab{} benchmark  \cite{zhai2019vtab}, which is a large-scale visual adaptation benchmark that consists of 19 visual classification tasks. \vtab{} can be further categorized into three groups: (1) Natural tasks, comprising images captured by standard cameras, (2) Specialized tasks, consisting of images captured by specialized equipment such as medical images, and (3) Structured tasks, which involve images containing geometric information (e.g., location, distance). Following \cite{jia2022vpt, zhang2022neural}, each task in the \vtab{} benchmark is split into 80\% train and 20\% val  (the latter is used for hyper-parameter tuning). The final model used for evaluation is trained using the full 1,000 examples in each dataset. Top-1 classification accuracy is used as the performance metric and is reported on a percentage scale.

\textbf{Implementation Details.}
For \vtab{} tasks, we follow \cite{jia2022vpt} to use the AdamW optimizer \cite{loshchilov2018fixing} with cosine learning rate decay to train the model for 100 epochs with a batch size of 64. The only trainable parameters are trainable prompts and the classification head. The data preprocessing strategies are also followed as \cite{jia2022vpt}. Following \cite{jia2022vpt, cheng2023e2vpt}, we search for the best tuning hyperparameters on val set for each task. 
The default prompt length, learning rate, weight decay, and $\lambda$ are set as 50, 0.001, 0.0001, and 0.95.

\begin{table}[htbp]
\centering
\caption{The comparison results against state-of-the-art methods on the \vtab{} benchmark. The best and second best results are highlighted in \textbf{bold} and \underline{underline}.}
\label{tab:vtab_compare}
    \begin{tabular}{lc | ccc|c}
    \toprule
     & \rotatebox{90}{\# param (M)} & \rotatebox{90}{Natural} & \rotatebox{90}{Specialized} & \rotatebox{90}{Structured} & \rotatebox{90}{Average} \\
    \midrule
    Full~\cite{jia2022vpt} & 85.8 & 74.5 & 83.4 & 47.7 & 68.5  \\
    Linear~\cite{jia2022vpt} & \textbf{0.04} & 69.1 & 77.1 & 26.9 & 57.9 \\ \midrule
    \multicolumn{4}{l}{\textcolor{gray}{Extra module-based methods:}} \\
    Adapter~\cite{houlsby2019parameter} & 0.16 & 79.0 & 84.1 & 58.5 & 73.9 \\
    LoRA~\cite{hu2022lora} & 0.29 & 79.5 & 84.6 & 60.5 & 74.9 \\
    NOAH~\cite{zhang2022neural} & 0.43 & 80.2 & 84.9 & \underline{61.3} & 75.5 \\
    SCT~\cite{zhao2023sct} & \underline{0.11} & \underline{82.2} & 85.9 & 59.9 & \underline{76.0} \\
    \midrule
    
    \multicolumn{4}{l}{\textcolor{gray}{Prompt-based methods:}} \\
    VPT~\cite{jia2022vpt} & 0.64 & 78.5 & 82.4 & 55.0 & 72.0 \\
    DAM-VP~\cite{huang2023diversity} & 2.52 & 81.3 & 83.8 & 54.3 & 73.1 \\
    \eevpt~\cite{cheng2023e2vpt} & 0.27 & 80.0 & 84.4 & 57.4 & 73.9 \\
    SA$^{2}$VP~\cite{pei2024sa2vp} & 0.25 & 80.8  & \underline{86.3} & 60.0 & 75.7 \\
    \rowcolor{mygray} SHIP (Ours) & 0.38 & \textbf{82.6} & \textbf{86.6} & \textbf{61.5} & \textbf{76.9} \\
    \bottomrule
    \end{tabular}
\end{table}

\subsection{Main Results}
We primarily compare our method with baseline methods across diverse visual recognition tasks. We conducted experiments on \vtab{}, initializing all methods with a ViT-B/16 \cite{vit} backbone pre-trained on ImageNet-21k \cite{imagenet}. 

% \textbf{Full-Shot Visual Recognition.} 
As depicted in Table \ref{tab:vtab_compare}, SHIP achieves superior overall performance across three \vtab{} tasks with 0.38M parameter budgets. For example, SHIP significantly outperforms prompt-tuning methods, surpassing VPT by 4.1\%, 4.2\%, and 6.4\% on natural, specialized, and structured tasks, respectively.
In addition, SHIP surpasses the state-of-the-art method \eevpt~\cite{cheng2023e2vpt} by a notable margin of $3.0\%$ in average accuracy of three tasks. 
The improvement in our performance can be attributed to SHIP's incorporation of discriminative prompt tokens into crucial semantic hierarchies, enhancing the extraction of task-related knowledge. Additionally, sharing prompts within each semantic level ensures sufficient feature aggregation.

\subsection{Ablation Study}\label{sec:ab}
\textbf{Effectiveness of Each Component.} 
We compare different components of our method when appended to the reproduced VPT-Deep baseline on two datasets. The ablation study presented in Table \ref{tab:ablation} indicates that each component significantly contributes to overall performance. SHIP surpasses the VPT-Deep baseline on Cifar100 \cite{cifar} and SUN397 \cite{sun397} by margins of 10.1\% and 11.2\%, respectively. The improvement with SIP is particularly notable, e.g., with a 7.7\% increase compared to the baseline on the Cifar100. This validates that our adaptive inference semantic-hierarchy strategy effectively extracts semantic representations, thereby enhancing performance. Additionally, due to the increase in sequence length from introducing SSP and AP, we compared it with directly adding an equivalent number of prompts in the SIP strategy. On Cifar100, this approach achieved an accuracy of 74.5\%, which is a slight improvement over the original accuracy of 74.2\% (the second row in Table \ref{tab:ablation}). This further validates the effectiveness of our method.

\begin{table}[htbp]
\centering
\caption{Ablation study on the Cifar100 and SUN397 datasets.}
% \vspace{-10pt}
\begin{tabular}{ccccc|cc}
\toprule
SIP & SSP & AP & PML & DA & Cifar100 & SUN397 \\ \midrule
& & & &  & 66.5 & 46.5 \\
\checkmark & & & & & 74.2 & 55.2 \\
\checkmark &  & & & \checkmark & 75.1 & 55.8 \\
\checkmark & \checkmark & & & \checkmark & 75.6 & 56.3 \\
\checkmark & \checkmark  & \checkmark  &  & \checkmark & 76.1 & 57.4 \\
\checkmark & \checkmark & \checkmark & \checkmark & \checkmark & \textbf{76.6} & \textbf{57.7} \\ \bottomrule
\end{tabular}
\label{tab:ablation}
\end{table}

\textbf{Effectiveness of Semantic Hierarchical Inference.} 
We initially evaluate the performance of various potential hierarchical prompting design choices as ablation of our proposed semantic hierarchy. These variants include uniform prompting and sharing prompts at specific positions and the results are presented in Table~\ref{tab:semantic_partition_comparison}.
Uniform prompting inserts prompts with a certain interval $I$. VPT-Deep can be seen as a particular case with $I=1$. Results show that it only achieves minor improvements over the baseline. As the interval increases, the performance decreases and eventually becomes comparable to VPT-Deep. This suggests that simply partitioning hierarchical stages uniformly may not be the optimal solution.
Furthermore, while sharing prompts show improvements at certain positions (e.g., 0→3), it still lags behind our proposed SHIP. Moreover, manually setting hierarchical intervals or sharing positions fails to account for inter-layer correlations, resulting in suboptimal performance. SHIP offers a superior solution by determining prompting hierarchies based on semantic-aware prior knowledge. This results in more obvious performance gains of 3.0\%.

\begin{figure}[htbp]
    \centering
    \begin{subfigure}{0.49\linewidth}
        \includegraphics[width=\linewidth]{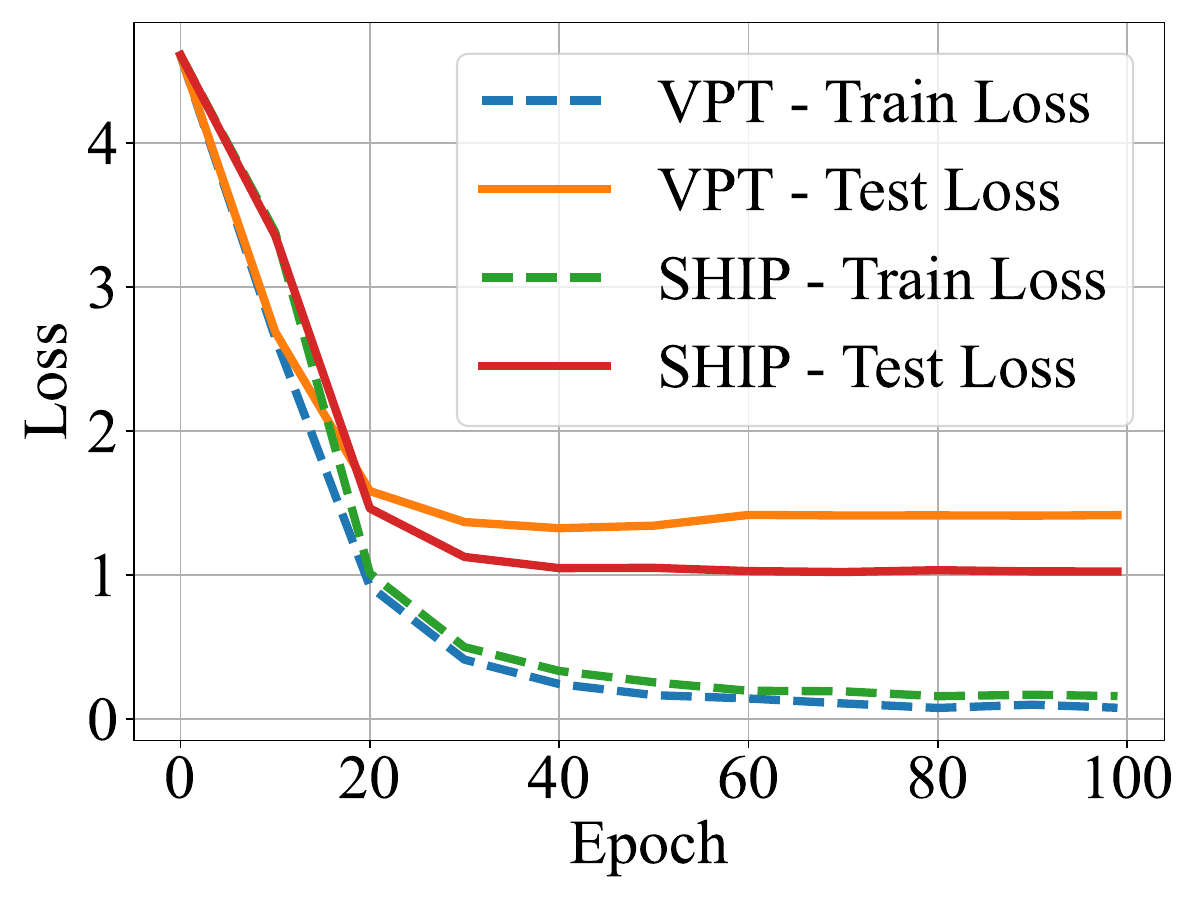} 
        \caption{Cifar100 \cite{cifar}}
        \label{fig:sub1}
    \end{subfigure}
    \hfill
    \begin{subfigure}{0.49\linewidth}
        \includegraphics[width=\linewidth]{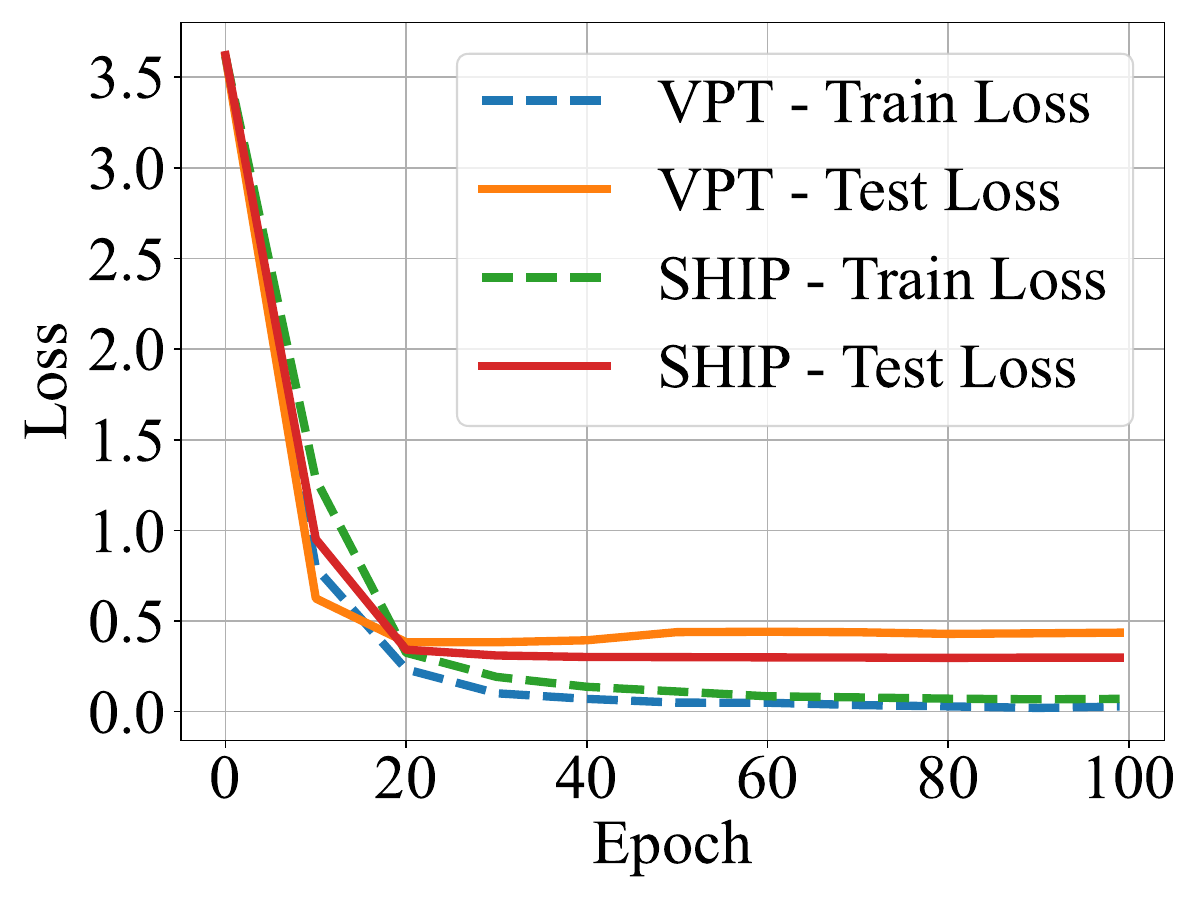} 
        \caption{Pets \cite{oxford_pets}}
        \label{fig:sub2}
    \end{subfigure}
    % \vspace{-10px}
    \caption{Loss comparison between VPT and SHIP.}
    \label{fig:training_logs}
\end{figure}

\begin{table}[htbp]
\caption{Comparison of various hierarchy partition strategies on the on \vtab{} \textit{Natural} tasks. Uniform prompting involves uniformly prompting across layers with $I$ intervals. ``Sharing $a$→$b$" indicates sharing prompts from $a$ to $b$ layers.}
% \vspace{-10px}
\centering
\begin{tabular}{lc | lc}
\toprule 
Method                        & Acc. &  Method     & Acc.  \\ \midrule
VPT-Deep                      & 77.8 &  + Sharing 0→3  &  79.7  \\  
+ Uniform prompting ($I$=2)   &  80.3    &  + Sharing 3→6  & 79.8 \\
+ Uniform prompting ($I$=3)   & 80.1    &  + Sharing 6→9  & 79.8 \\
+ Uniform prompting ($I$=4)   & 78.3     &  + Sharing 9→12  & 79.5 \\ \midrule
\rowcolor{gray!15}
 \multicolumn{4}{l}{+ Semantic independent prompting (Ours) \hspace{1.5cm} \textbf{80.8}}  \\  \bottomrule
\end{tabular}
\label{tab:semantic_partition_comparison}
% \vspace{-10px}
\end{table}

\textbf{Determination of Hyperparamter.} 
We compared several hyperparameters: the number of Attribute Prototypes \(K\), the layers with Attribute Prompts \(M_a\), the discriminative tokens \(N_m\) for matching loss, and the scale \(\lambda_d\) for Decoupled Attention. Results for different hyperparameters are presented in Table \ref{tab:ab_lambda}. Our findings indicate that SHIP achieves the best performance with \( K = 200 \) and fewer attribute prototypes lead to performance degradation. 
The optimal value for \( M_a \) is 2, with further increases \( M_a \) causing performance degradation. This is due to attribute aggregation with shallow tokens being inaccurate. 
While increasing \( N_m \) introduces more tokens for prompt matching, it can also bring in irrelevant background tokens, which hinder feature discrimination. 
For \( \lambda_d \), the optimal value is 0.1. Higher values substantially degrade performance by overly emphasizing prompt features and disrupting the pre-trained feature representations. Interestingly, without Decoupled Attention, the attention weight ratio between prompts and instance tokens also approximates 0.1, suggesting that \( \lambda_d = 0.1 \) aligns well with their relative balance. 
Additionally, the optimal hyperparameter values may vary across different datasets, and further tuning could potentially enhance performance.

\begin{table}[htbp]
\caption{Hyperparameters comparison on the Cifar100 dataset.}
% \vspace{-10px}
\centering
\begin{tabular}{cc|cc|cc|cc}
\toprule 
$K$ & Acc. & $M_a$ & Acc. & $N_m$ & Acc.   & $\lambda_d$           & Acc.  \\ \midrule
20    &  75.8  & 1     & 75.6   & 5     & 76.2  & 0.01 & 71.1 \\ 
50    &  76.2  & 2     & \textbf{76.6}    & 10 & \textbf{76.6}  & 0.1 & \textbf{76.6} \\
100   &  76.1  & 3     & 76.3   & 20    & 76.0   &  0.3 &72.8 \\
200   & \textbf{76.6}   & 4    & 75.8    & 50    & 75.6   & 0.5 & 40.5   \\
\bottomrule
\end{tabular}
\label{tab:ab_lambda}
% \vspace{-10px}
\end{table}

\textbf{Alleviate the Over-fitting.}
As we emphasized, SHIP can mitigate the degradation of pre-trained models and alleviate the overfitting that exists in VPT-Deep. Fig. \ref{fig:training_logs} illustrates the training and testing losses of both VPT and SHIP during the training process. We observed that VPT began to overfit around the $20$-th epoch, with its training loss continuing to decrease while the testing loss began to rise. In contrast, the testing loss of SHIP consistently decreased. 
This improvement can be attributed to SHIP's semantic hierarchical prompting strategies tailored for downstream tasks. The results demonstrate that SHIP effectively alleviates overfitting and achieves superior performance.

\section{Conclusion}
In this paper, we introduce a novel prompt-based fine-tuning method that improves pre-trained models by constructing semantic hierarchies and using semantic-specific prompts to optimize hierarchical features. We also incorporate attribute prompts for enhanced classification and propose prompt matching loss and decoupled attention to further boost performance and robustness. Overall, SHIP achieves significant performance gains compared to existing methods.

\section{Acknowledgments}
This work was supported by the National Natural Science Foundation of China under Grant 62072271.

\bibliographystyle{ieeetr}
\bibliography{strings,refs}

\end{document}